\documentclass[letterpaper]{article} 
\usepackage{aaai24}  
\usepackage{times}  
\usepackage{helvet}  
\usepackage{courier}  
\usepackage[hyphens]{url}  
\usepackage{graphicx} 
\usepackage{natbib}  
\usepackage{caption} 
\usepackage{algorithm}
\usepackage{algorithmic}

\usepackage{newfloat}
\usepackage{listings}
\DeclareCaptionStyle{ruled}{labelfont=normalfont,labelsep=colon,strut=off} 
\floatstyle{ruled}
\newfloat{listing}{tb}{lst}{}
\floatname{listing}{Listing}
\title{DSLE: A Learning Environment for Dark Souls Boss Encounters}
\author{
    Derin Gezgin\equalcontrib,
    Jim O'Connor\equalcontrib,
    Tanner Goodwin,
    Gary B. Parker
}
\affiliations{
    Autonomous Agent Learning Lab\\

    Connecticut College\\
    New London, Connecticut, USA\\
    \{dgezgin, joconno2, tgoodwin, parker\}@conncoll.edu
}

\usepackage{xspace}
\usepackage{pifont}
\usepackage{verbatim}
\usepackage{pgfplotstable}
\usepackage{booktabs}
\usepackage{float}
\usepackage{makecell}
\usepackage{amsmath}
\usepackage{amsfonts}
\usepackage{caption}
\usepackage{etoolbox}
\usepackage{dsfont}

\pgfplotsset{compat=1.18} 

\newcommand{\dsr}{Dark Souls: Remastered\xspace}
\newcommand{\dsapi}{DSAPI\xspace}
\newcommand{\dsle}{DSLE\xspace}
\newcommand{\cmark}{\ding{51}}
\newcommand{\xmark}{\ding{55}}

\newcommand{\symbolfootnote}[2]{%
  \begingroup
  \renewcommand{\thefootnote}{#1}%
  \footnote{#2}%
  \endgroup
}

\definecolor{codekeywords}{rgb}{0.10,0.20,0.65}
\definecolor{codestrings}{rgb}{0.55,0.20,0.10}
\definecolor{codecomment}{rgb}{0.20,0.45,0.25}
\definecolor{codelinenumbers}{rgb}{0.45,0.45,0.45}

\lstdefinestyle{dslepython}{
  language=Python,
  basicstyle={\scriptsize\ttfamily},
  keywordstyle=\bfseries\color{codekeywords},
  stringstyle=\color{codestrings},
  commentstyle=\itshape\color{codecomment},
  numberstyle=\scriptsize\color{codelinenumbers},
  showstringspaces=false,
  columns=fullflexible,
  keepspaces=true,
  tabsize=2,
  breaklines=true,
  numbers=left,
  xleftmargin=2em,
  aboveskip=2pt,
  belowskip=2pt
}

\newif\ifaddnewpage

\newcommand{\insertfig}{
  \setcounter{figure}{0}
  \includegraphics[width=\linewidth]{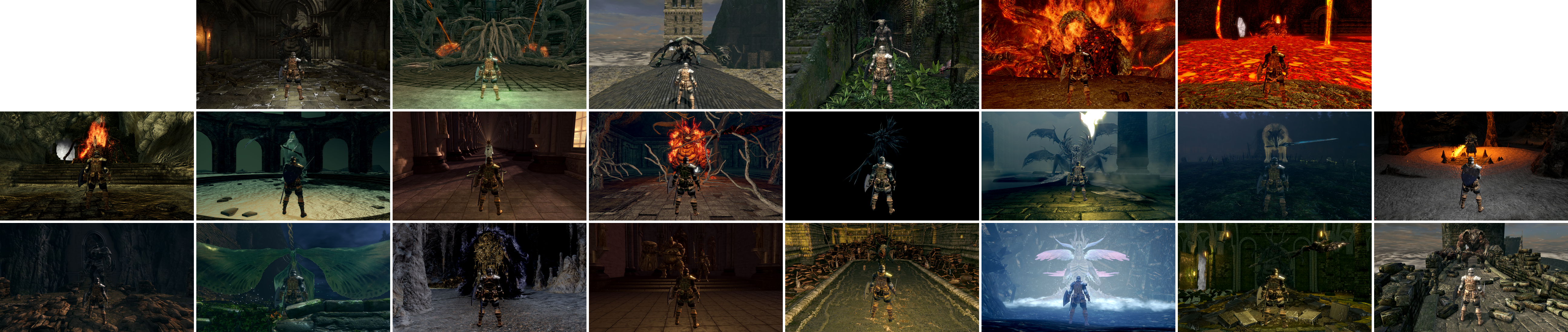}
  \captionof{figure}{
  Overview of the boss encounters from \textit{\dsr} supported in DSLE.
  These encounters span a diverse range of environments, enemy designs, and combat dynamics, providing a challenging benchmark for real-time control policies.
  }
  \label{fig:heroShots}
  \vspace{0.7em}
}

\makeatletter
\apptocmd{\@maketitle}{\centering\insertfig}{}{}
\makeatother

\begin{document}

\maketitle

\begin{abstract}
We introduce the Dark Souls Learning Environment (\dsle),
a containerized platform that presents all 22 boss encounters of \textit{\dsr} as game-playing agent benchmarks through a Gymnasium-style interface.
\dsle combines real-time combat, high-dimensional visual input, and sparse terminal rewards, with each environment step being a real action executed against the running game.
To support controlled comparison, we define \dsle-5,
a representative five-boss subset, spanning a melee fight, a spatially constrained arena, an environmental-hazard fight, a multi-target fight, and a fast final-boss fight,
that we recommend as the starting suite for agents built on \dsle.
On \dsle-5 we evaluate a random policy, an expert system, an evolutionary baseline, and PPO and DQN agents trained from visual input.
The expert system and the evolutionary baseline each defeat the Asylum Demon,
the game's tutorial boss (63\% and 43\% peak win rates), but none of the five methods defeats the other four \dsle-5 bosses;
PPO and DQN show no measurable learning ($\leq$0.33\% win rate on the tutorial boss, 0\% elsewhere) within a budget that already costs tens of wall-clock hours per run.
A broader study running the evolutionary baseline across all 22 encounters under
advantaged all level-50 stats yields wins on only a handful of additional
early-game bosses and leaves the rest unwon.
The failure cases range from sub-10-second deaths in cramped, multi-target encounters to minute-long stalemates that inflict almost no damage,
and we report them through survival time and damage dealt rather than win rate alone.%
\symbolfootnote{$\dagger$}{Code available at \texttt{github.com/ConnAALL/dsle}}

\end{abstract}

\section{Introduction}
\label{sec:intro}

Game playing has long served as a testbed for artificial intelligence,
providing controlled environments for evaluating decision-making under well-defined rules and objectives.
Early work focused on deterministic, fully observable domains such as chess and checkers \cite{shannon1950programming, CAMPBELL200257, 5392560},
and later on complex board games such as Go~\cite{silver_mastering_2017}, while more recent research has expanded to settings with stochasticity, partial observability, and high-dimensional observations.
A key development in this transition was the introduction of standardized evaluation environments.
The Atari Learning Environment (ALE) provides a diverse suite of games with a unified interface, enabling direct comparison across algorithms \cite{bellemare_arcade_2013}.
ALE has been widely used in both reinforcement learning and neuroevolution research, including Deep Q-Networks \cite{mnih_human-level_2015} and large-scale evolutionary approaches \cite{salimans_evolution_2017},
demonstrating that agents can learn directly from pixel input while highlighting challenges related to sample efficiency and exploration.

More recent environments have extended this paradigm to increasingly complex domains.
ViZDoom \cite{kempka_vizdoom_2016} introduces a three-dimensional partially observable setting,
while the StarCraft II Learning Environment \cite{vinyals_starcraft_2017} provides a large-scale real-time strategy domain with long horizons and hierarchical structure.
However, these platforms rely on purpose-built interfaces that expose structured state information.
Many modern commercial games do not provide such interfaces, limiting their use in reproducible evaluation pipelines.

\textit{Dark Souls: Remastered} presents one such domain.
The game is a third-person action role-playing game characterized by real-time combat, delayed and sparse rewards, and visually complex environments.
Progression is organized around boss encounters, each of which can be treated as a distinct task with unique dynamics, including differences in timing, spatial constraints, and interaction structure.
Prior work introduced the Dark Souls API (DSAPI), which enabled agent interaction through screen capture and computer vision \cite{oconnor_learning_2025}.
While this demonstrated the feasibility of learning from pixel input, the system relied on visual estimation of game state, required manual reset procedures, and supported only a single boss encounter.

In this work, we introduce the Dark Souls Learning Environment (DSLE),
a containerized system designed to support reproducible experimentation in \textit{\dsr}.
DSLE replaces vision-based state estimation with direct memory access, enabling reliable measurement of player and boss state.
The environment includes automated reset through save-state swapping, headless execution, and support for distributed training.
All 22 boss encounters are supported as independent tasks within a unified interface, each verified to load and reset correctly through the same pipeline.

To establish \dsle as a benchmark and to assess its difficulty,
we evaluate five representative boss encounters that we call \dsle-5 with a random policy, a scripted expert system, PPO and DQN agents trained from visual input, and Sparse Cosine Optimized Policy Evolution.
Only the tutorial boss, the Asylum Demon, is ever defeated: the expert system wins $63\%$ of episodes and SCOPE up to $43\%$ of its best generation, while PPO and DQN never exceed a $0.33\%$ win rate and show no measurable learning within a budget that already costs tens of wall-clock hours per run.
None of the five methods defeats any of the four harder \dsle-5 bosses, and this difficulty extends across the full 22-boss suite.

\section{Related Work}
\label{sec:rwork}

\subsection{Learning Environments for Games}

Standardized learning environments have been central to progress in reinforcement learning for games,
both by enabling controlled comparisons across algorithms and
by introducing new axes of difficulty that expose the limits of prior methods.
The Arcade Learning Environment is one of the first examples of a standardized learning environment for games,
supporting dozens of Atari 2600 games behind a common observation, action, and reward interface, enabling agents to be evaluated across diverse tasks without game-specific engineering \cite{bellemare_arcade_2013}.
This shared benchmark quickly became a standard testbed for both reinforcement learning \cite{mnih_human-level_2015} and black-box evolutionary search \cite{salimans_evolution_2017, such_deep_2018}.
Later environments retained the same interface-centric design while targeting properties Atari does not address.
ViZDoom renders a first-person, partially observable three-dimensional world from the raw screen buffer,
on the argument that the two-dimensional, third-person perspective of arcade games does not resemble real-world tasks \cite{kempka_vizdoom_2016},
and the StarCraft II Learning Environment adds imperfect information, a combinatorial action space, and long-horizon rewards \cite{vinyals_starcraft_2017}.
Following the same interface-centric design, MineRL applies it to the open world of \textit{Minecraft}, pairing a common agent interface with a large dataset of human demonstrations to study sample-efficient learning on long-horizon, hard-exploration tasks \cite{3367243.3367379}.
Extending this horizon further, the NetHack Learning Environment spans procedurally regenerated runs of tens of thousands of steps \cite{kuettler2020nethack},
while Procgen isolates generalization by scoring agents on held-out instances drawn from the same procedural distribution rather than the levels on which they trained \cite{cobbe_leveraging_2020}.
More recently, this benchmarking paradigm has extended to large language model agents, which are evaluated as game players across large collections of procedurally generated games through a common interface \cite{li_gvgai-llm_2025}.
Across this lineage, a purpose-built interface exposing structured state, rewards, and resets is what makes each environment reproducible as a benchmark.

A parallel line of work studies unmodified commercial games, but these systems are often difficult to reuse as general benchmarks.
AlphaStar achieved grandmaster-level StarCraft II play using league-based multi-agent training initialized from human replays \cite{vinyals_grandmaster_2019},
while OpenAI Five defeated world champions in Dota 2 through large-scale self-play \cite{openai_dota_2019}.
In games without such interfaces, agents have instead been trained from screen observations.
For example, behavioral cloning in Counter-Strike shows that a commercial first-person shooter can be controlled from pixels alone,
but through an offline imitation-learning pipeline rather than a live environment with automated resets \cite{pearce_counter-strike_2022}.
This leaves an important gap: a reproducible benchmark for a real-time, third-person commercial action role-playing game with close-range reactive combat, visual observations, automated resets, and repeatable episodes.
\dsle addresses this gap by exposing \textit{\dsr} as a standardized benchmark rather than a single-use integration.

\subsection{Prior Dark Souls Learning Systems}

The most directly related prior system is \dsapi \cite{oconnor_learning_2025}, a Python framework that exposes a limited agent interface to \textit{\dsr} through real-time screen capture, computer vision, and keyboard input emulation.
Using this interface, \citet{oconnor_learning_2025} trained NEAT agents to fight the Asylum Demon from raw visual input rather than an official game API.
Screenshots were downscaled to a compact RGB observation, and evolved policies selected from a discrete set of movement, attack, defensive, and healing actions.
The result was an important first demonstration that learning agents could interact with \textit{\dsr} through the same visual channel available to a player, and that evolved policies could discover useful
combat behavior in a commercial game without privileged simulator access.

In the published experiment, computer vision was used to extract player and enemy health from the game UI,
recognize menu and terminal-state elements, and support reset through a
predefined save file.
This made \textit{\dsr} accessible as a research target, but the demonstrated system was still organized around a single boss encounter and relied on visual estimation of game state rather than direct access to
structured environment variables.

\dsle builds on this line of work by changing the role of \textit{\dsr} from a single visually controlled experiment into a benchmark environment.
Instead of using computer vision as the primary state-estimation mechanism, \dsle combines rendered observations with direct process-memory instrumentation for reliable reward computation, termination checks, and diagnostic logging.
It also adds containerized execution, automated reset through scenario save states, isolated parallel instances, and boss-specific initialization procedures.
The resulting system preserves the central challenge of the earlier Dark Souls work, while providing a much wider task coverage and more structured execution infrastructure needed for systematic comparison.
Table~\ref{tab:dsapi_comparison} summarizes the main differences between the earlier \dsapi and \dsle.

\begin{table}[t]
\centering
\caption{Comparison between \dsapi by \citet{oconnor_learning_2025} and the Dark Souls Learning Environment (\dsle).}
\label{tab:dsapi_comparison}
\resizebox{\linewidth}{!}{
\begin{tabular}{lcc}
\hline
\textbf{Feature} & \textbf{\dsapi} & \textbf{\dsle} \\
\hline
Primary platform & Windows & Cross-platform (Docker) \\
Headless execution & \xmark & \cmark \\
Instances per machine & 1 & Multiple isolated instances \\
State extraction & Template matching & \makecell{Screen capture, template matching, \\ memory reading} \\
Boss coverage & 1 boss & All 22 bosses \\
Player configuration & Default setup & Scenario-specific save states \\
\hline
\end{tabular}}
\end{table}

\section{Dark Souls Learning Environment}
\label{sec:dsle}

\textit{\dsr} \cite{DarkSoulsRemastered2018} is an action role-playing game widely known for its challenging combat system and reputation for difficulty \cite{gaither_dark_2024}.
Player progression in the game depends on frequent boss fights, where the player must defeat powerful enemies with distinct attack patterns and behaviors.
These fights require sustained decision-making under strict timing constraints, as mistakes can result in significant damage, immediate defeat, or loss of progress.

From an AI perspective, boss fights provide a controlled but challenging setting for evaluating game-playing agents.
Each fight is a self-contained task with clear objectives, bounded spatial limits, and explicit termination conditions.
At the same time, the combat system demands reactive control, spatial awareness, and temporal coordination between offensive and defensive actions.
These properties make boss encounters in \textit{\dsr} a useful benchmark for studying adaptive control policies in real-time game environments.

The Dark Souls Learning Environment (\dsle) provides a scalable benchmark and software interface for training and evaluating agents in these encounters.
Rather than exposing the full game campaign, which includes long-horizon navigation, strategic decisions, item progression, and many smaller fights with non-player characters, \dsle isolates boss fights as repeatable combat scenarios.
Each task begins from a preconfigured save state that places the agent in the boss encounter, gives the agent direct control of the player character, and ends when the boss is defeated, the player dies, or a step limit is reached.
This focus keeps the task definition compact enough for controlled experimentation while retaining the core difficulty of \textit{\dsr}.

\subsection{Available Boss Encounters}
\label{subsec:bosses}

Each \dsle task is specified by four components: a boss identifier, a save state, a boss-specific initialization procedure, and an episode interface.
The save state fixes the player build, inventory, equipment, and location.
The standard \dsle-5 protocol reuses a single low-level starting character across all five encounters so that scores are comparable across bosses of very different intended difficulty,
while the exploratory 22-boss study in Section~\ref{subsec:beyond} uses one uniform advantaged build for every encounter and is reported separately for that reason.
Users can create additional scenario save files to study alternate character configurations, equipment loadouts, and inventory contents.
These custom saves make it possible to vary the difficulty of an encounter, for example by changing health, stamina, armor, weapon damage, or starting position while retaining the same control interface.
Scenario saves can also be paired with alternate game and display configurations, including resolutions up to $3840 \times 2160$.
The initialization procedure loads the save, reaches the boss entrance, performs any encounter-specific setup, and then transfers control to the agent.
After control is transferred, the same observation, action, reward, and termination interface is used across bosses.

\dsle supports all 22 boss encounters in \textit{\dsr}.
We verified that every encounter can be loaded, reset, and stepped through the interface so any of them can serve as an evaluation task, although the learning experiments in this paper use the \dsle-5 subset defined in Section~\ref{sec:experiments}.
These encounters differ in spatial layout, number of enemies, vulnerability windows, phase structure, environmental hazards, and the ideal strategy to beat them.
A single method can perform well on short reactive melee fights while failing on bosses that require waiting, target switching, long-horizon positioning, or distanced attacks.

For instance, the \textit{Asylum Demon}, the game's opening boss, is designed to be beaten on a first attempt: it signals every slow hammer swing and only requires basic dodging and attacking.
The \textit{Capra Demon} is fought in what players widely consider the smallest arena in the game and is accompanied by two fast dogs that rush the player the instant the fight begins, making opponent selection and positioning crucial.
Other encounters move the difficulty onto entirely different axes.
\textit{Chaos Witch Quelaag} continuously floods her arena with lava that lingers on the ground and shrinks the safe areas over time;
\textit{Ornstein and Smough} is a two-on-one fight in which defeating either boss heals and empowers the other, punishing the intuitive strategy of splitting damage;
and the \textit{Bed of Chaos} is not a regular fight at all, but a fight in which the player must destroy environmental targets while the floor collapses into instantly lethal pits.
All bosses in \textit{\dsr} require different skills, from reactive timing to hazard avoidance, target prioritization, and multi-stage planning, allowing \dsle to function as a benchmark with a built-in difficulty gradient rather than a single repeated challenge.
Table~\ref{tab:boss_characteristics} summarizes the full supported boss set, listing the design and the control skills each encounter emphasizes.

\definecolor{tagMeleeColor}{RGB}{180,90,52}
\definecolor{tagPositionColor}{RGB}{0,114,178}
\definecolor{tagMultiColor}{RGB}{0,158,115}
\definecolor{tagPhaseColor}{RGB}{204,121,167}
\definecolor{tagRangedColor}{RGB}{86,180,233}
\definecolor{tagHazardColor}{RGB}{213,94,0}
\definecolor{tagPerceptionColor}{RGB}{0,139,139}
\definecolor{tagObjectiveColor}{RGB}{230,159,0}
\definecolor{tagTransferColor}{RGB}{90,90,90}
\newcommand{\benchTag}[2]{%
  \begingroup
  \setlength{\fboxsep}{1pt}%
  \colorbox{#1!12}{\textcolor{#1!72!black}{\scriptsize\bfseries #2}}%
  \endgroup\hspace{1pt}%
}
\newcommand{\tagMelee}{\benchTag{tagMeleeColor}{Melee}}
\newcommand{\tagPosition}{\benchTag{tagPositionColor}{Positioning}}
\newcommand{\tagMulti}{\benchTag{tagMultiColor}{Multi-target}}
\newcommand{\tagPhase}{\benchTag{tagPhaseColor}{Phase/state}}
\newcommand{\tagRanged}{\benchTag{tagRangedColor}{Ranged}}
\newcommand{\tagHazards}{\benchTag{tagHazardColor}{Hazards}}
\newcommand{\tagPerception}{\benchTag{tagPerceptionColor}{Perception}}
\newcommand{\tagObjective}{\benchTag{tagObjectiveColor}{Objective}}
\newcommand{\tagTransfer}{\benchTag{tagTransferColor}{Transfer}}

\begin{table*}[t]
\centering
\footnotesize
\caption{The 22 supported \dsle boss encounters.
The ``Characteristics'' column summarizes each encounter's design, and ``Benchmark Focus'' uses a fixed, color-coded tag set to mark the control skills and pressures each boss emphasizes:
\tagMelee{}: reactive close combat;
\tagPosition{}: spatial positioning;
\tagMulti{}: multiple simultaneous targets;
\tagPhase{}: phase or state transitions;
\tagRanged{}: ranged or projectile threats;
\tagHazards{}: environmental hazards;
\tagPerception{}: hard-to-perceive targets;
\tagObjective{}: objective completion beyond pure combat;
and \tagTransfer{}: reuse of behavior learned on a related boss.}
\label{tab:boss_characteristics}
\setlength{\tabcolsep}{2pt}
\begin{tabular}{>{\raggedright\arraybackslash}p{0.16\textwidth}p{0.62\textwidth}p{0.21\textwidth}}
\toprule
\textbf{Boss} & \textbf{Characteristics} & \textbf{Benchmark Focus} \\
\midrule
Asylum Demon &
Tutorial boss with slow, heavily telegraphed attacks and a simple arena. &
\tagMelee \tagPosition\\
Bed of Chaos &
Puzzle boss focused on destroying environmental targets while the arena collapses. &
\tagObjective \tagHazards \tagPhase\\
Bell Gargoyles &
Rooftop boss where a second gargoyle joins mid-fight. &
\tagMulti \tagPhase\\
Capra Demon &
Early boss fought in a cramped arena with two aggressive dogs and limited room. &
\tagMulti \tagPosition\\
Ceaseless Discharge &
Large boss fought near lava, with long-reaching attacks and narrow attack windows. &
\tagRanged \tagHazards\\
Centipede Demon &
Lava-area boss with long-reaching attacks, restricted safe ground, and severable limbs. &
\tagHazards \tagPosition\\
Chaos Witch Quelaag &
Area denial boss with sword attacks, lava pools, and explosive fire area attacks. &
\tagHazards \tagMelee\\
Crossbreed Priscilla &
An invisible boss that is tracked mainly through footprints in the snow. &
\tagPerception \tagPosition\\
Dark Sun Gwyndolin &
Ranged magic boss fought down a long corridor with repeated projectile attacks. &
\tagRanged \tagPosition\\
Demon Firesage &
Late-game Asylum-style demon with heavy melee swings and explosive leap attacks. &
\tagTransfer \tagHazards \tagMelee\\
Four Kings &
Abyss boss where additional Kings spawn over time, creating escalating pressure. &
\tagMulti \tagPhase\\
Gaping Dragon &
Large boss with slow, telegraphed slams, charges, and long recovery windows. &
\tagPosition \tagMelee\\
Gravelord Nito &
Boss supported by numerous skeleton enemies, with Nito relying on slow melee attacks. &
\tagMulti \tagPosition \tagMelee\\
Great Grey Wolf Sif &
Large sword-wielding wolf boss fought in an open arena with wide sweeping attacks. &
\tagMelee \tagPosition\\
Gwyn, Lord of Cinder &
Final boss with fast, aggressive sword attacks and high pressure at close range. &
\tagMelee \tagPosition\\
Iron Golem &
Large fortress boss fought on a narrow high platform where falls are a major threat. &
\tagHazards \tagPosition \tagMelee\\
Moonlight Butterfly &
Flying ranged boss that attacks with magic projectiles before landing briefly. &
\tagRanged \tagPosition\\
Ornstein and Smough &
Dual-boss dual-phase fight with a fast spear user and a slow heavy attacker. &
\tagMulti \tagPhase \\
Pinwheel &
Weak boss that teleports, casts fire magic, and creates clones of itself. &
\tagPerception \tagMulti\\
Seath the Scaleless &
Large crystal puzzle boss with sweeping attacks, and an initial vulnerability mechanic. &
\tagObjective \tagPosition \tagPhase\\
Stray Demon &
Stronger Asylum Demon variant with heavy melee attacks and large magic explosions. &
\tagTransfer \tagMelee \tagHazards\\
Taurus Demon &
Early boss encounter. A large minotaur-like melee enemy fought on a narrow bridge. &
\tagMelee \tagPosition \tagHazards\\
\bottomrule
\end{tabular}
\end{table*}

\subsection{Environment Interface}
\label{subsec:env}

\dsle follows a Gymnasium-style interface \cite{towers_gymnasium_2024}:
\begin{equation}
    o_{t+1}, r_t, \mathrm{terminated}, \mathrm{truncated}, \mathrm{info}
    = \mathrm{step}(a_t).
\end{equation}
Here $a_t$ is the discrete action selected by the agent, $o_{t+1}$ is the next observation, $r_t$ is the scalar reward,
$\mathrm{terminated}$ and $\mathrm{truncated}$ are Boolean episode flags, and $\mathrm{info}$ is a dictionary for logging and evaluation.
Listing~\ref{lst:dsle_python_example} shows a minimal single-episode Python interaction loop for a boss instance, and Table~\ref{tab:dsle_interface} summarizes the default observation, action, termination, and reward configuration.

\begin{listing}[t]
\begin{lstlisting}[style=dslepython]
import dsle

env = dsle.make(
  boss="asylum_demon",
  instance="dsr-1",
  obs_mode="grayscale",
  max_steps=7200,
)

try:
  obs, info = env.reset()
  ret = 0.0
  done = False
  while not done:
    action = env.action_space.sample()
    obs, reward, term, trunc, info = env.step(action)
    ret += reward
    done = term or trunc
finally:
  env.close()
\end{lstlisting}
\caption{Minimal single-episode interaction loop for a \dsle boss encounter.}
\label{lst:dsle_python_example}
\end{listing}

\begin{table}[t]
\centering
\caption{\dsle interface summary for the default observation, action, termination, and reward configuration.}
\label{tab:dsle_interface}
\resizebox{\linewidth}{!}{
\begin{tabular}{ll}
\toprule
\textbf{Component} & \textbf{Interface} \\
\midrule
Default observation & Grayscale frame, $1 \times 600 \times 800$, \texttt{uint8} \\
Alternative observations & RGB frame, $3 \times 600 \times 800$; diagnostic \texttt{state\_only} mode \\
Action space & \texttt{Discrete(14)} player-control actions \\
Default action hold & 250 ms per selected action \\
Default step limit & 7200 steps per episode \\
Termination & Boss defeated or player death \\
Truncation & Step limit reached \\
Default reward terms & Boss damage, player damage, step penalty, win bonus, death penalty \\
\bottomrule
\end{tabular}}
\end{table}

\begin{figure}[t]
    \centering
    \includegraphics[width=\columnwidth]{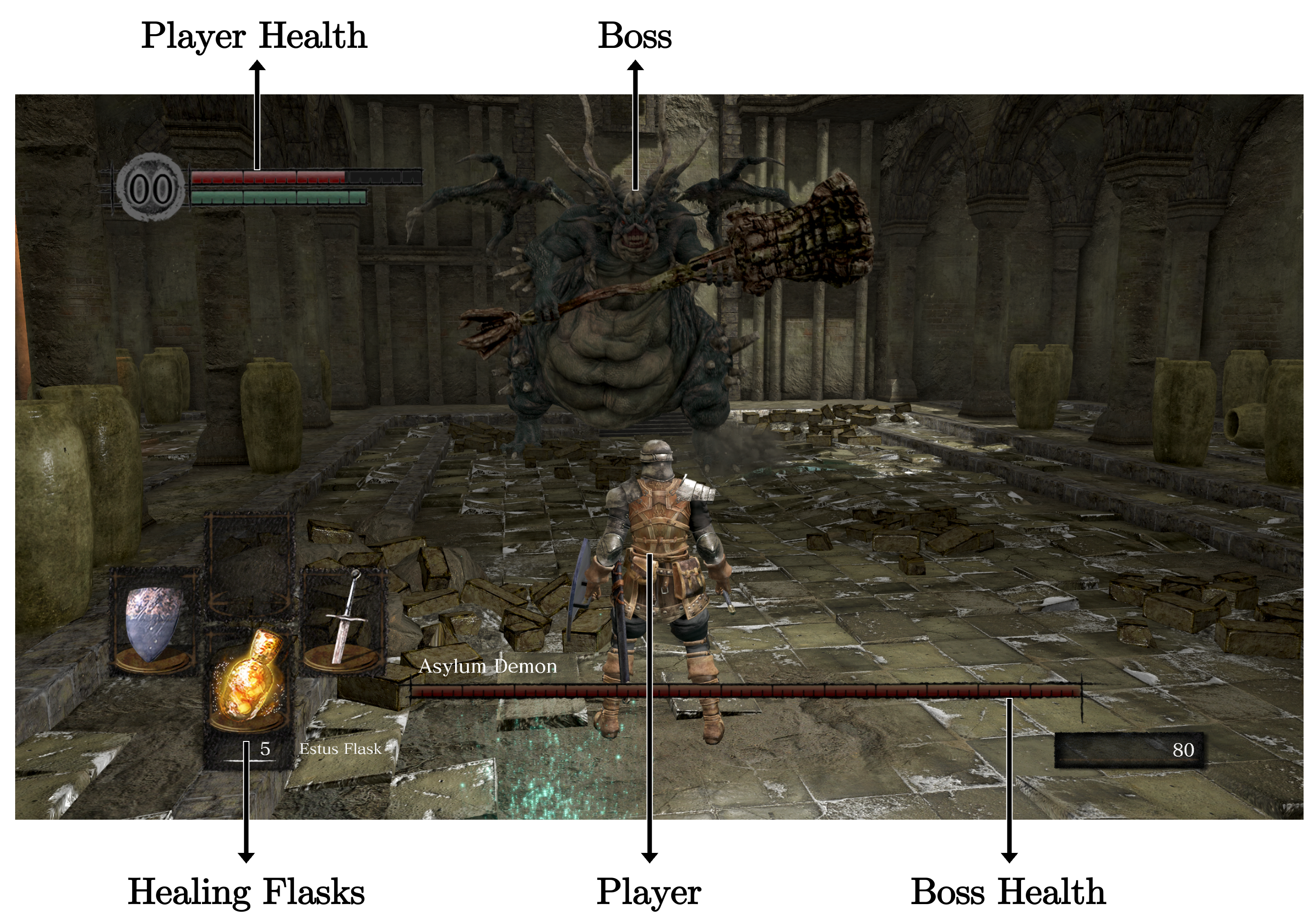}
    \caption{In-game frame from \textit{\dsr} showing the player-view observation used by \dsle, including the arena, the player, the boss (in this case, Asylum Demon), and the game HUD with the player and boss health bars.}
    \label{fig:dsle_ingame_frame}
\end{figure}

\paragraph{Policy Input Space.}
The default policy input is a captured game frame from the player's view.
In the standard visual mode, \dsle converts this frame to a single-channel grayscale tensor with observation space \texttt{Box(0, 255, (1, 600, 800), uint8)}.
This image includes the same rendered arena, player character, boss, animation state, camera perspective, and game HUD that are visible in the running game, as shown in Figure~\ref{fig:dsle_ingame_frame}.
The environment also includes preprocessing wrappers for resizing observations, stacking frames, clipping rewards, changing time limits, and recording episodes.
The environment records metrics such as player health, boss health, death count, lock-on state to the boss, and whether the boss has been defeated.
These measurements are used for reward computation, termination checks, and logging, and are provided to the user in the \texttt{info} dictionary.

\paragraph{Policy Output Space.}
The agent outputs one integer from \texttt{Discrete(14)}.
The 14 actions are four directional moves (forward, backward, left, and right), a light attack, a strong attack, a heal, a backstep, four directional rolls (forward, backward, left, and right), and two raw mouse-click actions (left- and right-click) for parrying.
Each selected action is injected into the game and held for a configured duration, 250 ms by default, before the next action is queried; the agent therefore issues roughly four decisions per second of wall-clock game time.
This fixed hold makes the control rate explicit and reproducible rather than tying it to the rendering frame rate.
Lock-on to the boss is handled automatically by the environment at a fixed interval when enabled, as maintaining camera focus is a low-level bookkeeping action rather than a strategic decision in most boss encounters.
Keeping it automatic leaves the 14 actions focused on movement, attacks, healing, and dodging, while the agent still has to learn combat timing and positioning.
Automatic lock-on is configurable and can be disabled, and the current lock-on state is read from memory and logged for every step.

\paragraph{Environment Outputs.}
After executing $a_t$, \dsle returns the next observation $o_{t+1}$ in the selected observation space, a scalar reward $r_t$, two Boolean episode flags, and an \texttt{info} dictionary.
The \texttt{terminated} flag is true when the boss is defeated or the player dies.
The \texttt{truncated} flag is true when the episode reaches the step limit.
The reset and step metadata include diagnostic fields such as the boss identifier, instance name, current step, raw player and boss health, damage dealt and taken during the step, and whether the episode ended in a win.

\subsection{Stochasticity and Evaluation Protocol}
\label{subsec:stochasticity}

\begin{figure}[t]
\centering
\includegraphics[width=\columnwidth]{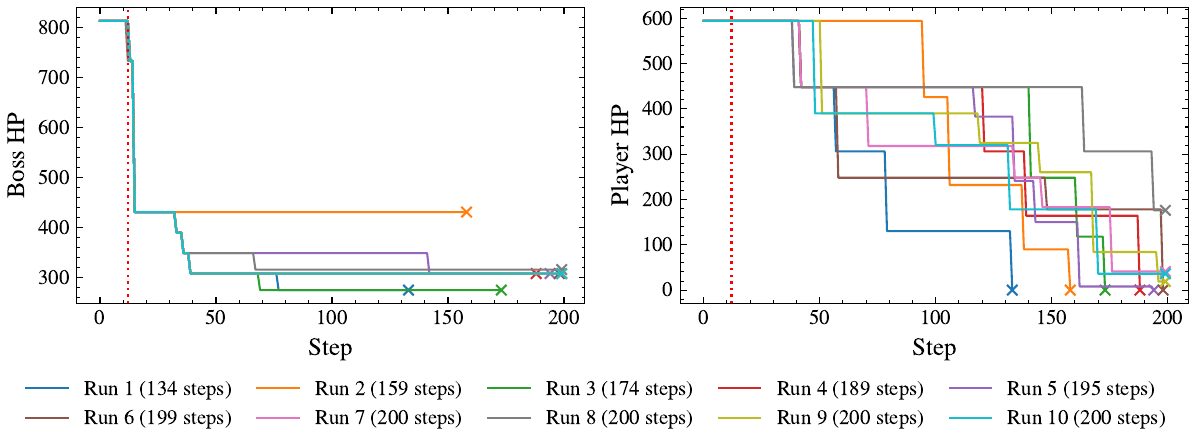}
\caption{Asylum Demon stochasticity replay test.
Ten runs begin from the same save state and execute the same set of actions.
The two panels show boss-health and player-health trajectories.
Despite identical replayed inputs, the health trajectories diverge by step 12.}
\label{fig:asylum_demon_stochasticity}
\end{figure}

Although every episode begins from a fixed save state, \dsle should not be treated as a fully deterministic environment.
To measure this directly, we ran a test with Asylum Demon in which each trial executed the same set of actions: attack, move forward, attack, and roll forward, repeated for up to 200 steps.
The test logged the selected action, boss health, and player health after each step, while leaving the game process, timing, animation system, and boss behavior unseeded by the environment.
Across ten trials, the trajectories first diverged at step 12 and terminated after 134--200 steps.
Figure~\ref{fig:asylum_demon_stochasticity} shows the resulting boss-health and player-health trajectories for these 10 runs.
An agent can exploit early regularities in a fight such as the boss entrance sequence, but reliable performance still requires feedback from the current episode state.

\subsection{Runtime Architecture}
\label{subsec:runtime}

Running \textit{\dsr} as an RL environment requires an orchestration layer over the game.
\dsle executes the unmodified Windows game inside an Ubuntu-based Docker container through Wine.
Rendering is provided through a virtual X11 display, and DXVK translates DirectX calls to Vulkan for GPU-backed execution.
Each game instance receives its own display number, VNC port, Wine prefix, save directory, desktop name, and XDG runtime directory.
This isolation avoids collisions between Wine processes and allows several independent encounters to run in a single container on a single machine.

Episode resets in the environment are automated.
Before an episode, \dsle copies the target scenario save into the active game save directory, navigates the game's menus using template matching, loads the character, waits for the boss-entry prompt, and enters the arena.
Some encounters require additional setup operations, such as walking to a trigger, waiting for a template, or skipping a cutscene.
These operations are part of the environment reset procedure and are performed before the agent-controlled combat phase.

The action layer sends keyboard and mouse events through X11 input injection.
From the game's perspective, the agent is pressing the same keys and mouse buttons as a human player.
The observation layer captures frames from the virtual display.
\dsle combines visual templates with process-level instrumentation of the running game process.
At runtime, \dsle locates the game process associated with the active instance and opens the Linux \texttt{/proc/<pid>/mem} interface for reading the memory state.
The reader resolves fixed base pointers for the player, boss, event-flag, lock-on, and menu structures, then follows boss-specific pointer chains declared in the configuration files.
Through these memory reads, \dsle retrieves player health, maximum player health, death count, boss health, boss-defeated flags, and lock-on state.
This instrumentation is used to compute rewards, recover from menus, return to the title screen, and verify terminal states.

\subsection{Runtime Scalability and Reproducibility}
\label{subsec:scalability}

In \dsle, each environment step is bounded by the real execution of the game and the selected action-hold duration. Consequently, the environment is not intended to match the throughput of JAX-native environments~\cite{matthews2024craftax, brax2021github}.
This creates a trade-off between single-instance throughput and the fidelity of running an unmodified commercial game.
\dsle is designed to be scalable through parallelism rather than high single-instance throughput.
It can generate multiple game instances, assign them separate displays and Wine prefixes, and run them concurrently inside the same container or across multiple machines.

\dsle runs on any Linux host with an NVIDIA GPU and Docker.
Each concurrent game instance adds roughly 0.3~GB of GPU memory and 2--3~GB of system RAM on top of a shared $\sim$1.2~GB GPU baseline,
measured at the default $600 \times 800$ observation size, and a single game installation occupies about 30~GB of disk that is shared across all instances on a machine.
Because system RAM rather than GPU memory is the binding constraint at scale, the number of concurrent instances is set by host memory:
our experiments ran five instances per machine, but we were able to run up to 30 instances on a system with an NVIDIA RTX 5090 (32~GB of VRAM) and 128~GB of RAM,
where the instances together used roughly 10~GB of VRAM and about 80~GB of RAM.
The variation is driven almost entirely by episode resets rather than by the steps themselves: encounters in which the agent dies within a few seconds produce thousands of short episodes,
and each reset must copy a save file, navigate the menus, and re-enter the arena.

For reproducibility, \dsle includes Docker build files, setup scripts, scenario save states, environment checks, evaluation scripts, and baseline agents.
The container can also expose VNC sessions and episode recordings, allowing researchers to inspect live agent behavior directly instead of relying only on scalar logs.
The game itself is not redistributed with the code: users provide their own legally obtained copy of \textit{\dsr},
while the open-source artifact supplies the environment wrapper, automation, and benchmark code around that dependency.
This follows the same distribution model as other game-based environments that ship only the research interface and require the user to supply the game,
such as the StarCraft~II Learning Environment, which requires a separately installed copy of the game~\cite{vinyals_starcraft_2017}.
A single user-provided installation is reused across an entire instance pool, so scaling to many parallel environments does not require a separate purchased copy per running instance.

\section{Experiments}
\label{sec:experiments}

\dsle is a benchmark environment for all 22 bosses supported by \textit{\dsr}.
Our experiments aim to characterize environment stochasticity rather than treat it as a fixed open-loop puzzle,
provide non-learning reference points through random and expert-system controllers,
and establish learning baselines from both reinforcement learning and neuroevolution.
We evaluate five baselines on a representative subset of five bosses that we call \dsle-5: a random policy, a scripted expert system, and PPO and DQN agents under a single shared protocol,
together with the neuroevolutionary policy.

The \dsle-5 subset is chosen to cover the distinct pressures of the benchmark: it includes a tutorial melee fight (Asylum Demon), a spatially constrained arena (Capra Demon), an environmental-hazard fight (Chaos Witch Quelaag), a multi-target fight (Ornstein and Smough), and a fast final-boss fight (Gwyn, Lord of Cinder).
We recommend \dsle-5 as the starting suite for anyone building an agent on \dsle.
Running all 22 bosses is costly in both wall-clock time and compute, so this representative subset gives researchers a lower-cost target on which to develop and compare methods before committing to the full set.
\dsle-5 is not the only possible suite that can be constructed from the 22 bosses. The
encounter tags in Table~\ref{tab:boss_characteristics} can be used to construct further boss suites with specific characteristics.

\subsection{Evaluation Metrics and Episode Protocol}

Each episode begins by loading a boss-specific scenario save, navigating into the encounter, and transferring control to the agent at the start of combat.
The policy observes grayscale visual frames and selects from the shared \texttt{Discrete(14)} action space described in Section~\ref{sec:dsle}.
An episode terminates when the boss is defeated or the player dies, and it is truncated when the configured step limit is reached.
Because the game process has its own internal randomness, agent seeds control only agent-side randomness such as action sampling, parameter initialization, and any other random initializations throughout the pipeline.
Setting the seeds in agent scripts does not make the game dynamics deterministic, as we demonstrate in Section~\ref{subsec:stochasticity}.

\subsection{Random and Expert-System Baselines}

We evaluate two non-learning baselines on \dsle-5: a random policy and a rule-based expert system.
These baselines provide non-learning reference points: uniform random action selection and a simple hand-written reactive policy.

The random policy samples uniformly from the action space defined in Section~\ref{sec:dsle} at every step. The expert-system baseline is a small rule-based controller.
It observes the diagnostic state exposed in the \texttt{info} dictionary, heals when player health falls below 40\% of the maximum player health observed so far, waits through a short heal cooldown, and otherwise alternates between moving forward and attack.
This controller is not meant to approximate expert human play; it is a simple reactive combat policy for the benchmark.
After the scenario setup procedure, the boss is initialized in front of the agent, so the simplest hand-written combat prior is to move forward, attack, and spend heals only when health is low.
This tests whether a minimal closed-loop combat heuristic is sufficient for any encounter while revealing how quickly the benchmark becomes harder when timing, positioning, target management, or hazard avoidance matter.

\subsection{Reinforcement Learning Baselines}

We train PPO~\cite{schulman_proximal_2017} and DQN~\cite{mnih_human-level_2015} baselines on \dsle-5 using visual observations only.
The policies never receive the memory-extracted game-state scalars; those are used solely for reward computation, termination, and logging.
Both agents observe the grayscale $(1,600,800)$ frame, which the preprocessing pipeline downsamples to $(1,84,84)$ before it is passed to the convolutional architecture introduced by \citet{mnih_human-level_2015}, consisting of three convolutional layers followed by a fully connected layer.
For PPO, the network has a shared convolutional trunk with separate policy and value heads.
For DQN, the network predicts action values and uses epsilon-greedy exploration.

The reinforcement learning agents receive the shaped environment reward
\begin{equation}
r_t = -0.001 + \Delta h_b - 0.25\Delta h_p
      + 100 \cdot \mathds{1}_{\mathrm{win}} - 10 \cdot \mathds{1}_{\mathrm{death}},
\end{equation}
where $\Delta h_b$ is the fraction of boss health removed during the step and $\Delta h_p$ is the fraction of player health lost during the step.
This reward provides intermediate feedback for damage dealt and damage received while preserving the terminal win bonus as the main success signal.

We use a single fixed evaluation protocol for both methods and all five bosses, with no per-boss or per-method tuning of the experimental setup: each method and boss is run with the same five agent-side seeds, the same target budget of 100{,}000 environment steps per run, and the same network architecture and reward.
The protocol is fixed across bosses; PPO and DQN differ only in their algorithm-specific hyperparameters, as described below.
This keeps the comparison reproducible and avoids tailoring the protocol to any individual encounter.
PPO uses rollouts of 128 steps, four optimization epochs, minibatches of 32, learning rate $2.5\times 10^{-4}$ with annealing, $\gamma=0.99$, generalized advantage estimation with $\lambda=0.95$, clipping coefficient 0.1, and entropy coefficient 0.01.
DQN uses a replay buffer of 100{,}000 transitions, minibatches of 32, learning rate $10^{-4}$, $\gamma=0.99$, epsilon annealed from 1.0 to 0.01 over the first 10\% of training, learning starts after 10{,}000 steps, and target-network updates every 1000 steps.

\subsection{Neuroevolutionary Baseline}
\label{subsec:neuro}

We also include a neuroevolutionary baseline: Sparse Cosine Optimized Policy Evolution (SCOPE),
a lightweight derivative-free policy for high-dimensional visual game inputs~\cite{10.1609/aiide.v21i1.36834, 10.1007/978-3-032-15635-8_14}.
SCOPE is motivated by the fact that evolutionary methods become harder to optimize as visual inputs increase the number of free policy parameters.
It addresses this by applying a two-dimensional discrete cosine transform to each grayscale observation,
retaining the low-frequency and high-energy coefficient block,
sparsifying low-magnitude coefficients,
and mapping the resulting sparse coefficient matrix to action preferences with a bilinear affine policy.
We optimize this compact policy with CMA-ES~\cite{hansen_completely_2001}.
Five independent SCOPE trials are reported for each boss in \dsle-5.

SCOPE runs use grayscale $600\times800$ observations, with the SCOPE policy used to select one of the 14 policy actions exposed by the runtime.
SCOPE uses truncation parameter $k=100$, sparsification parameter $p=90$, CMA-ES initial step size $\sigma=0.5$, and no explicit population-size override.
With 14 output actions, this gives a $1{,}514$-parameter chromosome.
The default CMA-ES population under this configuration contains 25 candidates, so each 40-generation training trial evaluates 1{,}000 candidate episodes.
Candidate evaluations are parallelized across Docker/Wine game instances locally, which significantly reduces the wall-clock time required to train a single SCOPE policy.

SCOPE uses the fitness
\begin{equation}
f = (H_p + (1 - H_b)) \times 100 ,
\end{equation}
where $H_p$ is the final player-health ratio and $H_b$ is the final boss-health ratio.
A perfect episode in which the player defeats the boss without taking damage obtains a fitness of 200.
Because fitness adds the survival and damage terms, an intermediate value does not by itself indicate a win:
an agent can score highly by either removing most of the boss's health or retaining high player health, even without a kill.
Wins are therefore recorded separately, from the boss-death flag rather than from fitness.

\subsection{Computational Cost}
\label{subsec:cost}

Because every environment step executes a real action against a running copy of the game, the computational overhead of \dsle is limited by the game's execution speed.
At the default 250~ms action hold the agent makes roughly four decisions per second of game time.
The 100{,}000-step budget is close to seven hours of game time per run before a single reset is counted.
Episodes end on a win or a death and truncate at 7{,}200 steps, and because episode length is determined by how long the agent survives rather than by a fixed schedule,
the number of episodes inside that budget varies by boss and by method:
pooled over five seeds, PPO completes between $1{,}684$ and $6{,}423$ episodes per boss and DQN between $2{,}284$ and $12{,}549$, as shown in Table~\ref{tab:winrate}.
A single run takes $9$--$21$ hours for PPO and $12$--$41$ hours for DQN.
This cost can be reduced through parallelism by running five game instances per machine across several machines.

\section{Results}
\label{sec:results}

In this section, we present results primarily from \dsle-5.
Table~\ref{tab:winrate} reports the per-episode win rate of all five baselines on \dsle-5.
On the tutorial boss, Asylum Demon, the methods separate cleanly: the scripted expert wins $63\%$ of episodes, SCOPE wins $21\%$, and PPO and DQN win under $0.4\%$, while the random policy never wins.
At its best generation, SCOPE reaches a $43\%$ win rate, which exceeds the $35\%$ reported at the same best-generation granularity for the NEAT agent of \citet{oconnor_learning_2025},
the only prior neuroevolutionary result on \textit{\dsr}; this makes SCOPE the strongest neuroevolutionary policy reported on this boss to date.
For the four harder bosses, every method records zero wins across thousands of episodes.

In Section~\ref{subsec:rl}, we show that the tested gradient-based RL baselines show no sign of learning within a budget that already costs tens of wall-clock hours per run.
They are in fact the weakest family we tested, beaten on an early boss by a simple derivative-free evolutionary policy and an expert system.
In Section~\ref{subsec:failuremodes}, we show that agents fail in distinct regimes, and these failure modes can be inferred from survival time and boss damage.
Finally, in Section~\ref{subsec:beyond}, we look beyond \dsle-5 to the rest of the benchmark's 22 bosses, and find that the difficulty extends well past the five we studied in depth.

\begin{table}[t]
\centering
\small
\setlength{\tabcolsep}{3pt}
\begin{tabular}{lccccc}
\toprule
 & \multicolumn{5}{c}{Wins / episodes} \\
\cmidrule(lr){2-6}
Method & \makecell{Asylum\\\footnotesize Tut.} & \makecell{Capra\\\footnotesize Early} & \makecell{Quelaag\\\footnotesize Mid} & \makecell{O\&S\\\footnotesize Hard} & \makecell{Gwyn\\\footnotesize Hard} \\
\midrule
Random          & $0/100$   & $0/100$   & $0/100$  & $0/100$  & $0/100$ \\
Expert          & $63/100$  & $0/100$   & $0/100$  & $0/100$  & $0/100$ \\
PPO             & $5/1971$  & $0/6423$  & $0/1684$ & $0/5309$ & $0/5892$ \\
DQN             & $8/2469$  & $0/12549$ & $0/2284$ & $0/8203$ & $0/9942$ \\
SCOPE           & $1065/5000$ & $0/5000$ & $0/5000$ & $0/5000$ & $0/5000$ \\
\bottomrule
\end{tabular}
\caption{Per-episode results on \dsle-5 under the standard save states, reported as wins\,/\,episodes to make sample sizes explicit.
Random and Expert are evaluated for $100$ episodes per boss.
PPO and DQN are pooled over five seeds trained for $100{,}000$ steps each, and SCOPE is pooled over five seeds of $40$ generations with $25$ candidates each ($5{,}000$ episodes per boss).
Only the tutorial boss is ever beaten; all four harder bosses hold every method to zero wins.}
\label{tab:winrate}
\end{table}

\subsection{Reinforcement Learning Baselines Cannot Learn Within Budget}
\label{subsec:rl}

\begin{figure*}[t]
\centering
\includegraphics[width=\textwidth]{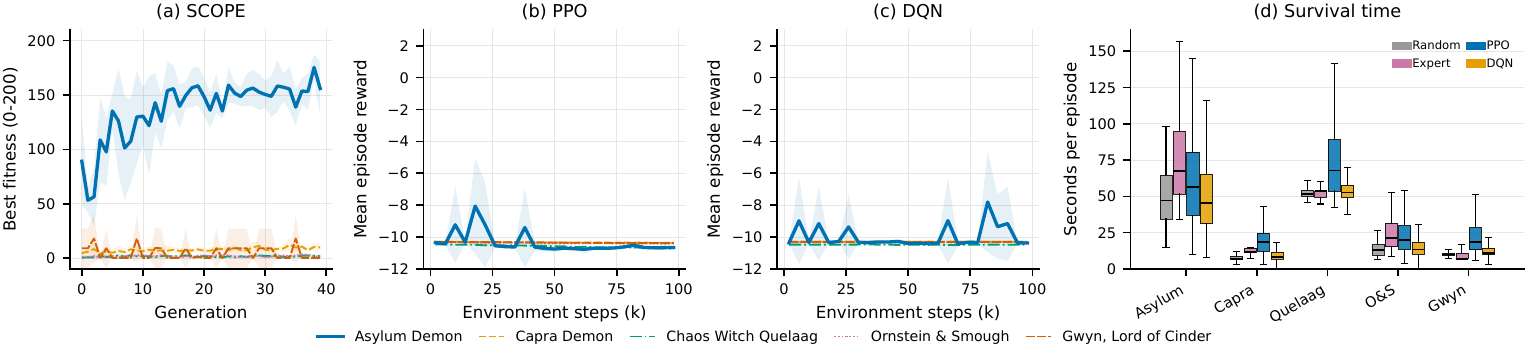}
\caption{
Learning curves and failure modes on \dsle-5 (five seeds; mean $\pm$ std in (a)--(c)).
\textbf{(a)} SCOPE best-of-generation fitness value outlined in Section~\ref{subsec:neuro}. 
\textbf{(b,c)} PPO and DQN mean episode reward stay pinned at the death floor ($\approx-10$) on every boss for the full $100$k-step budget; the boss curves overlap because no method makes measurable progress.
\textbf{(d)} Per-episode survival time by baseline (boxes span the IQR, whiskers $1.5\times$IQR):
agents die within seconds in the cramped Capra and fast Gwyn fights but survive longest on Quelaag while dealing almost no damage, so long survival is not progress.
}
\label{fig:learningcurves}
\end{figure*} 

PPO and DQN show no measurable learning on any boss within the $100{,}000$-step budget.
Figure~\ref{fig:learningcurves}(b,c) plots episode reward against environment steps. Every curve is pinned at the death floor near $-10$ and never trends upward.
As a win contributes $+100$ and this is never reached, the only signal is a small shaped-damage term set against a $-10$ death penalty, leaving a near-flat, sparse reward landscape.
A single run takes $9$--$21$ hours for PPO and $12$--$41$ hours for DQN (Capra Demon, whose short episodes force constant save-state reloads, is the most expensive),
i.e. tens of wall-clock hours per run across the $50$-run matrix.
The evolutionary baseline does better only on the tutorial:
SCOPE's best-of-generation fitness (Figure~\ref{fig:learningcurves}(a)) climbs steadily on Asylum but stays flat near the floor on all four harder bosses, 
and it registers kills only on Asylum.
With the given compute budget, gradient-based RL methods were only able to defeat the tutorial boss,
and did so less frequently than both SCOPE and the expert-system agents.

\subsection{How Agents Fail}
\label{subsec:failuremodes}

The failures split into two regimes, both visible in how long an agent survives (Figure~\ref{fig:learningcurves}(d)) and how much boss health it removes.
In the small-arena boss encounters the agent dies almost immediately: median survival is $7$--$19$\,s on Capra and $7$--$19$\,s on Gwyn across methods.
This is very little time to land meaningful hits, so even the expert system removes only $\sim\!7$ out of $771$ boss HP on Capra Demon.
In the environmental-hazard fight the opposite happens: on Quelaag the agents survive \emph{longest} of all (medians $52$--$68$\,s) yet still deal almost no damage,
because the boss approaches slowly and the agent is killed by lava and ranged attacks before the agent engages.

\subsection{Difficulty Beyond \dsle-5}
\label{subsec:beyond}

The evaluation above characterizes \dsle-5, but \dsle supports all 22 encounters.
To understand how far the difficulty extends, we ran an exploratory study over the full set,
asking not how well an agent scores but whether each boss can be won \emph{at all} under conditions that favor the agent.
In \dsr, a character's overall strength is summarized by its stats: 
the total of the points invested across attributes such as Vitality, Endurance, and Strength.
While the experiments above use the game's low-level starting character,
here we evaluated SCOPE on every encounter under advantaged save states built at all level-50 stats,
which give the player substantially more health, stamina, and weapon damage.

These advantaged saves are intentionally easier than the standard save states used in Section~\ref{sec:experiments} and Table~\ref{tab:winrate};
the aim is not comparability with those results but a lower bound on which encounters are solvable under any configuration.
Under this eased setting, SCOPE won only a handful of early-game encounters:
$96\%$ of episodes on the Capra Demon, $88\%$ on the Taurus Demon, $16\%$ on Pinwheel, and $8\%$ on the Bell Gargoyles;
while the large majority of the 22 bosses remained unwon.
Compared to Table~\ref{tab:winrate}, SCOPE wins most advantaged episodes in Capra Demon,
showing that per-encounter difficulty in \dsle can be tuned through the save state
while the observation, action, and reward interface is held fixed.

Several encounters outside \dsle-5 are not harder versions of the Asylum-style melee task that can be solved with better equipment.
The Bed of Chaos has a single unit of health and cannot be beaten by dealing damage,
the player must destroy two environmental targets while the arena floor collapses into instantly lethal pits.
Seath the Scaleless is invulnerable until the player turns away from the boss to find and destroy a separate crystal that regenerates on each attempt.
None of these fights is a reactive melee task;
they demand objective completion, target switching, multi-stage planning, and adaptation to non-stationary dynamics.

\section{Conclusion}
\label{sec:conclusion}

We introduced the Dark Souls Learning Environment (\dsle), a containerized benchmark for game-playing agents in \textit{\dsr}.
\dsle exposes all 22 of the game's boss encounters through a single Gymnasium-style interface, runs the unmodified game inside Docker over Wine,
and resets each encounter automatically from a scenario save state so that experiments run headlessly and in parallel without manual intervention.
The encounters span a wide range of control problems, from a tutorial boss to multi-target, hazardous, and objective-based fights that demand qualitatively different skills.
By bringing standardized benchmarking to a real-time commercial action RPG,
\dsle opens an evaluation setting whose domain and deployment constraints are not covered by existing learning environments.

Our baselines establish that \textit{\dsr} has a multi-tiered difficulty structure, with the harder environments being unsolved.
An expert controller wins the majority of its Asylum Demon episodes but removes less than 1\% of the health of every harder \dsle-5 boss;
the evolutionary SCOPE policy wins only the tutorial Asylum Demon; and PPO and DQN show no measurable learning within a budget that already costs tens of wall-clock hours per run.
Across all five baselines, the mid- and hard-tier bosses go unwon and agents fail in interpretable, encounter-specific ways.
Reading survival time alongside the fraction of boss health removed, we find fights that kill agents within seconds,
fights that agents survive but cannot meaningfully damage, and fights whose two-on-one or multi-stage structure defeats any fixed reactive behavior.

\subsection{Future Work}
\label{subsec:future}

\dsle opens the door to a broader set of baselines and agent families because each can be evaluated through the same interface, save-state protocol, and boss suite.
The most immediate direction is methods built for sparse, long-horizon, real-time control~\cite{lu2025intelligent}.
The flat learning curves of our model-free baselines point toward agents that can exploit denser intermediate structure,
whether through reward shaping, recurrent or transformer-based policies that track combat state across frames,
model-based rollouts that amortize the high per-step cost of a real game~\cite{schrittwieser_mastering_2020, Hafner2025},
or imitation from human or scripted demonstrations~\cite{3367243.3367379, 3600270.3602059, aytar2018playing} to overcome the sparse success signal.

The structured layout of \dsle also allows future work in curriculum and transfer learning.
Because the encounters form a graded sequence and share several mechanics, such as the Asylum Demon and its later,
tougher variants, agents could be trained from easy to hard rather than on each boss in isolation, and skills learned on one fight could transfer to related ones.
The difficulty of any encounter can be tuned by changing the player's health, stamina, equipment,
or starting position while keeping the interface fixed, which supports both curriculum design and controlled studies of robustness.
Because \dsle presents many distinct tasks under one interface, it is also a natural setting for multi-task and continual learning.
Finally, as most of the architecture of \dsle is game-agnostic, the same methodology can be applied to other souls or soulslike games. 

\subsection{Limitations}
\label{subsec:limit}

\dsle executes a real action against the real game at a fixed control rate, so single-instance throughput is far below that of emulator or simulator-based benchmarks, and a single training run can take tens of hours.
We address this through parallelism: instances are isolated and run concurrently within a container and across machines, and a single game installation is shared across an entire pool of instances.
This makes wall-clock cost a function of how many instances a researcher can run, but it does not make \dsle a high-throughput environment, and methods that need hundreds of millions of steps will find it expensive.
The cost is also uneven across bosses, since encounters that end in fast deaths spend proportionally more time on resets than on agent control.

Using \dsle requires a legally obtained copy of \textit{\dsr},
which the open-source artifact treats as an external dependency in the same way that the StarCraft~II Learning Environment relies on a separately installed game client~\cite{vinyals_starcraft_2017}.
A single purchased copy is reused across all instances on a machine, so scaling does not require buying additional copies.

All the baselines were run on the five-boss \dsle-5 subset under a single fixed protocol of five seeds and a 100{,}000-step budget,
chosen so that the comparison is reproducible and free of per-boss tuning rather than to chase the best possible score.
These results bound what current methods achieve at a realistic compute budget;
they do not establish the asymptotic difficulty of any encounter, and they leave most of the 22 bosses, as well as memory-based, model-based, and curriculum-driven approaches, open for future work.

\bibliography{references}

\end{document}